\documentclass{article}

\usepackage{iclr2027_conference,times}
\usepackage{latexsym}
\usepackage[T1]{fontenc}
\usepackage[utf8]{inputenc}
\usepackage{microtype}
\usepackage{inconsolata}
\usepackage{graphicx}
\usepackage{subcaption}
\usepackage{wrapfig}
\usepackage{amsmath}
\usepackage{amssymb} 
\usepackage{graphicx}
\usepackage{multirow}
\usepackage{tabularx}
\usepackage{booktabs}
\usepackage{enumitem}
\usepackage{xspace}
\usepackage[table,svgnames]{xcolor}
\usepackage[normalem]{ulem}
\usepackage{hyperref}
\usepackage{url}

\newcommand{\MethodName}{ROSD\xspace}

\title{Beyond Imitation: Reflective On-Policy Self-Distillation for LLM Reasoning
}

\author{Ziqi Zhao\textsuperscript{1},
Xinyu Ma\textsuperscript{2},
Liu Yang\textsuperscript{1},
Yujie Feng\textsuperscript{1},
Daiting Shi\textsuperscript{2},
Jingzhou He\textsuperscript{2},
\\
\textbf{Xin Xin\textsuperscript{3}},
\textbf{Zhaochun Ren\textsuperscript{4}},
\textbf{Xiao-Ming Wu\textsuperscript{1}}
\\
\textsuperscript{1}The Hong Kong Polytechnic University
\quad
\textsuperscript{2}Baidu Inc.
\\
\textsuperscript{3}Shandong University
\quad
\textsuperscript{4}Leiden University
\\
}

\iclrfinalcopy

\begin{document}
\maketitle
\begin{abstract}

On-policy self-distillation (OPSD) improves the reasoning capabilities of large language models (LLMs) by providing dense token-level supervision for on-policy rollouts. However, existing OPSD methods often yield limited gains on complex reasoning tasks and suffer from severe training instability. We identify two key causes: conditioning the self-teacher on a complete verified solution encourages imitation of complete reference trajectories rather than extraction of transferable reasoning insights, while indiscriminate full-response distillation imposes superfluous supervision on already-valid reasoning prefixes. Together, these issues suppress reasoning diversity and contribute to late-stage mode collapse.

We propose Reflective On-policy Self-Distillation (\MethodName), which distills transferable reasoning insights rather than complete reference trajectories. For each erroneous rollout, a self-reflector contrasts it with a correct rollout from the same group to derive a corrective idea and identify the sentence containing the first reasoning error. The corrective idea provides the self-teacher with targeted guidance, while the diagnosed error boundary allows \MethodName to mask out the distillation loss over the valid prefix and apply token-level distillation only from the first erroneous sentence onward. Experiments across multiple reasoning benchmarks and model backbones show that \MethodName consistently outperforms standard OPSD and reinforcement learning baselines, better preserves reasoning diversity, stabilizes training, and mitigates late-stage mode collapse. Code is available at \url{https://github.com/ZiqiZhao1/ROSD}.

\end{abstract}

\section{Introduction}\label{sec:introduction}



Post-training plays a key role in improving the reasoning ability of large language models~\citep{achiam2023gpt,kumar2025llm}.
Among post-training approaches, reinforcement learning with verifiable rewards (RLVR) has emerged as an effective paradigm for optimizing models in verifiable domains~\citep{shao2024deepseekmath,yu2026dapo,team2025kimi}.
However, typical RLVR algorithms such as GRPO~\citep{guo2025deepseek,shao2024deepseekmath} rely on outcome rewards to compute response-level advantages for model optimization.
As a result, they provide no token-level supervision over intermediate reasoning steps, which limits their ability to further improve reasoning quality.
OPSD addresses this limitation by using the model itself as a teacher, additionally conditioned on a verified correct solution~\citep{hubotter2026reinforcement,zhao2026self,shenfeld2026self,he2026self,li2026unifying}. It provides dense supervision by minimizing the token-wise KL divergence between the student and self-teacher distributions along on-policy rollouts.



\begin{figure*}[t]
\centering
\includegraphics[width=0.203\textwidth]{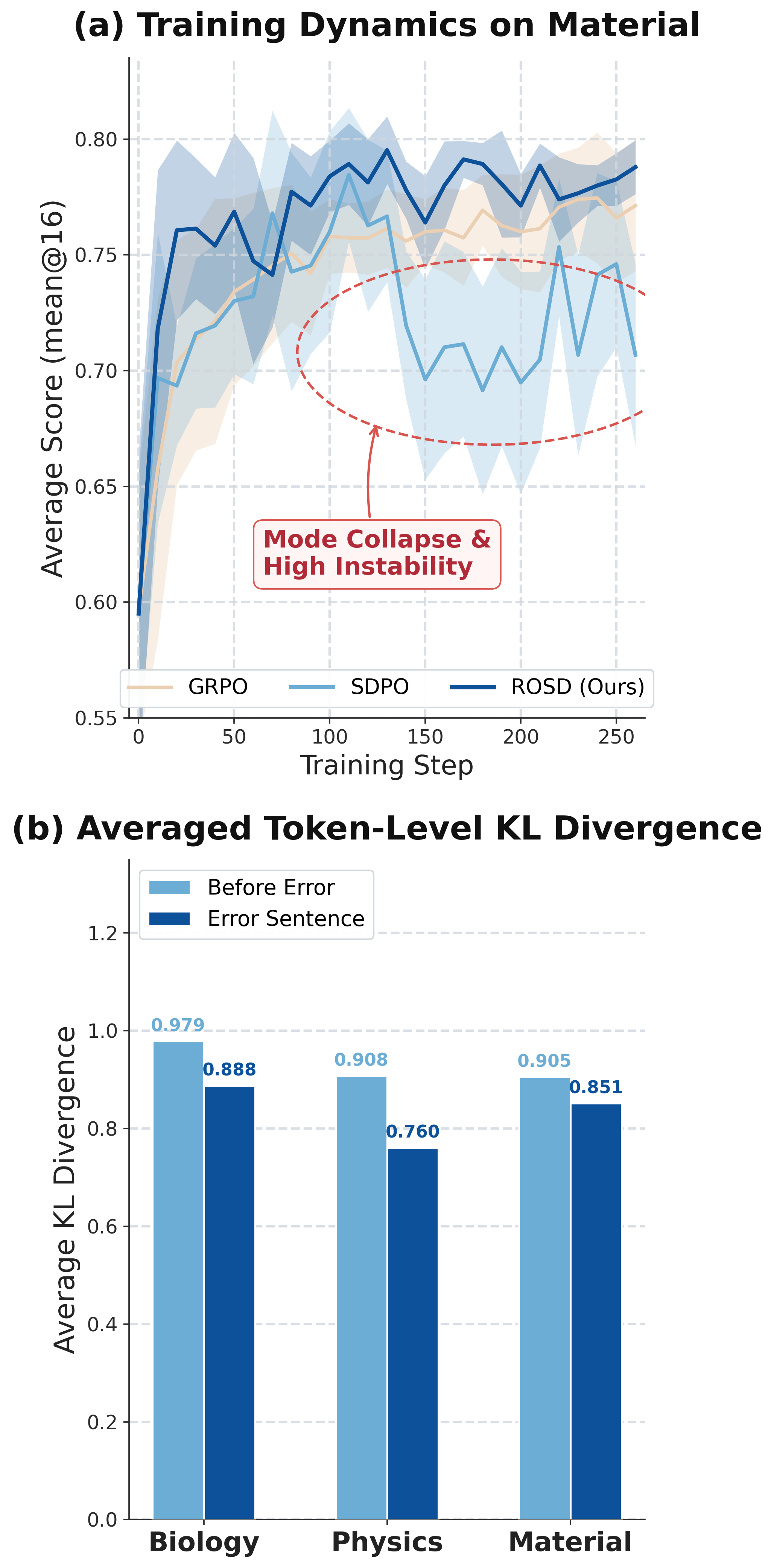}\hfill
\includegraphics[width=0.795\textwidth,trim=0 19bp 0 0,clip]{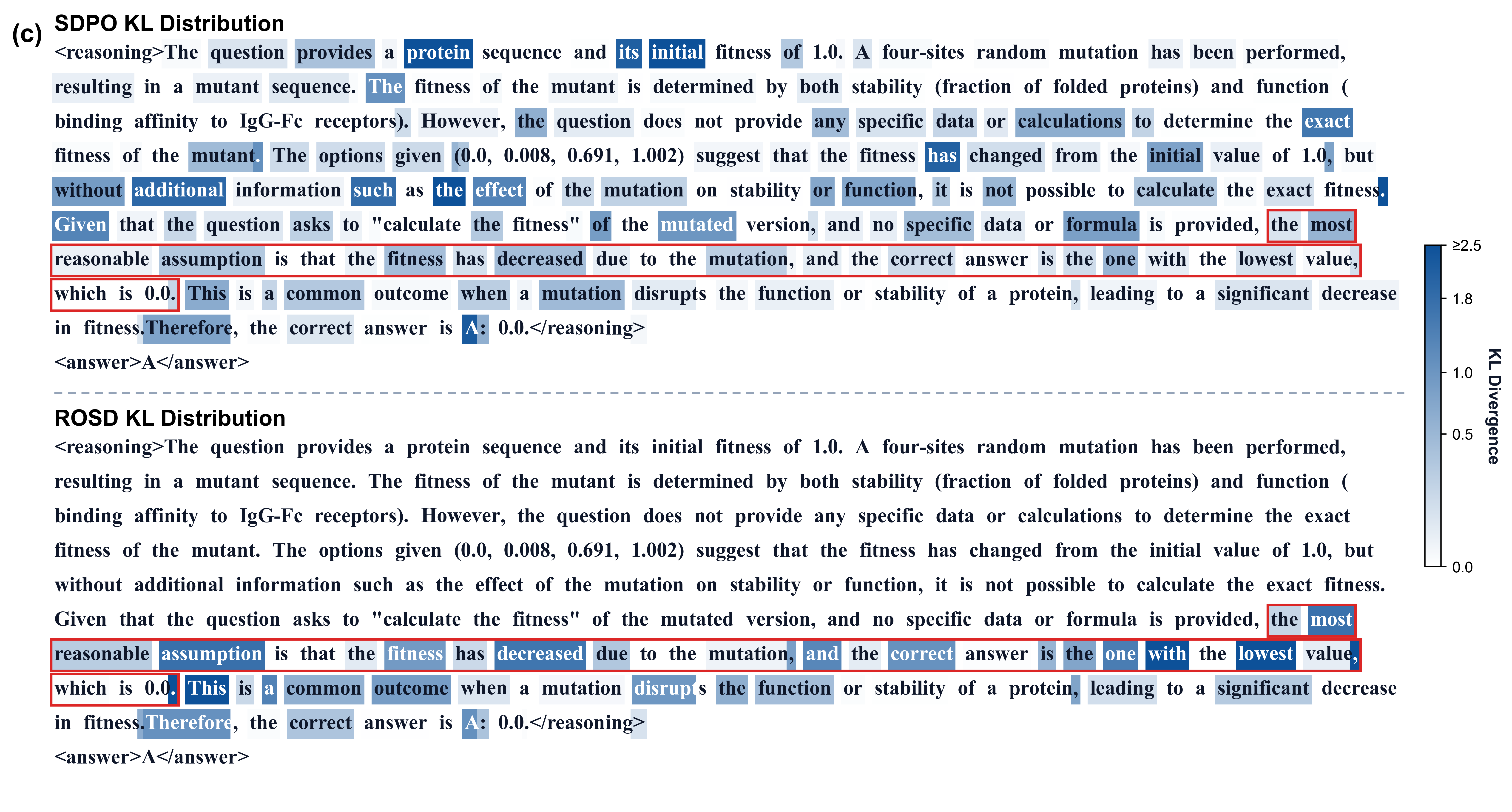}
\caption{
\textbf{(a)} Evaluation score dynamics during training on the Material benchmark.
\textbf{(b)} Average per-token KL divergence induced by SDPO across three scientific reasoning benchmarks. The valid reasoning prefix (\textit{Before Error}) consistently receives greater distillation pressure than the sentence containing the first reasoning error (\textit{Error Sentence}).
\textbf{(c)} Token-level KL divergence on a representative Biology rollout under SDPO (top) and \MethodName (bottom). Darker colors indicate larger KL divergence, and the red rectangle marks the first diagnosed reasoning error. SDPO imposes substantial distillation pressure throughout the valid prefix while applying comparatively weak supervision to the error. In contrast, \MethodName masks out the KL loss over the valid prefix and applies token-level distillation only from the diagnosed error onward. This example provides a token-level illustration of the aggregate pattern observed in \textbf{(b)} and demonstrates how \MethodName preserves valid reasoning while focusing its corrective signal on the erroneous suffix.
}
\label{fig:overview}
\vspace{-0.4cm}
\end{figure*}

Despite the promise of dense supervision, standard OPSD often exhibits training instability and suboptimal reasoning performance. As shown in Figure~\ref{fig:overview}(a), a representative OPSD method, SDPO~\citep{hubotter2026reinforcement}, undergoes pronounced performance degradation and mode collapse during the later stages of training. We hypothesize that this behavior stems from indiscriminate trajectory-wide imitation. By conditioning the self-teacher on a complete verified solution and applying distillation throughout the response, standard OPSD aligns the student with the solution-conditioned teacher even when the student's intermediate reasoning is already valid. Since many problems admit multiple valid reasoning paths, such alignment may penalize an alternative yet sound strategy simply because it differs from the reference solution. Over time, this can suppress reasoning diversity and bias the model toward a narrow set of solution patterns.


To substantiate this hypothesis, we analyze the token-level KL divergence induced by SDPO on erroneous rollouts. For each rollout, we identify the valid prefix preceding the first logical error and the sentence containing that error. As shown in Figure~\ref{fig:overview}(b), we compare the average per-token KL divergence over these two regions across three scientific reasoning benchmarks. Strikingly, the distillation pressure on the valid prefix is consistently greater than that on the first erroneous sentence. Since the prefix is already logically valid, such strong pressure does not directly contribute to correcting the identified error. Figure~\ref{fig:overview}(c) illustrates this aggregate finding with a representative Biology rollout: SDPO imposes substantial KL penalties throughout the valid prefix while applying comparatively weak supervision to the first erroneous sentence. This case study illustrates how the solution-conditioned self-teacher can induce superfluous imitation of the reference trajectory, including its specific phrasing, rather than provide targeted supervision for the actual reasoning failure. By penalizing valid alternative derivations, such redundant supervision may suppress reasoning diversity and concentrate the policy on a narrow set of brittle solution patterns, thereby contributing to mode collapse. We provide the complete analysis protocol and case study, including the problem, rollout, and error diagnosis, in Appendices~\ref{app:token_kl_analysis} and~\ref{app:case_study_biology}.

Taken together, these findings suggest that self-distillation should focus on \emph{transferable reasoning insights} rather than enforce full-trajectory imitation. Motivated by this principle, we propose Reflective On-policy Self-Distillation (\MethodName), a framework that replaces direct imitation of complete correct solutions with reflection-guided, error-localized distillation. As illustrated in Figure~\ref{fig:method_overview}(c), for each erroneous rollout, a self-reflector contrasts it with the shortest correct rollout from the same group to derive a corrective idea and identify the sentence containing the first reasoning error. The corrective idea conditions the self-teacher on what went wrong and how to repair it, providing targeted guidance without exposing the teacher directly to the complete correct trajectory. The diagnosed error boundary further allows \MethodName to preserve the valid reasoning prefix by masking out its distillation loss and applying token-level distillation only from the first erroneous sentence onward. For a correct rollout, the self-reflector instead summarizes the key idea underlying its successful reasoning, which serves as the auxiliary context for self-teaching. In this way, \MethodName distills concise, rollout-specific reasoning guidance while preserving valid alternative derivations, thereby stabilizing training and mitigating late-stage mode collapse.

Our primary contributions are summarized as follows:
\begin{itemize}[leftmargin=*,itemsep=1pt,topsep=2pt,parsep=1pt,partopsep=1pt]
    \item We empirically identify that standard OPSD's reliance on full-trajectory imitation and indiscriminate response-level distillation imposes superfluous supervision on valid reasoning prefixes, suppressing reasoning diversity and inducing training instability and late-stage mode collapse.

    \item We propose \MethodName, a reflection-guided OPSD framework that distills transferable reasoning insights rather than complete reference trajectories, enabling targeted reasoning correction while preserving valid prefixes.

    \item Extensive experiments across multiple challenging reasoning benchmarks and model backbones show that \MethodName consistently outperforms standard OPSD and RLVR baselines, better preserves reasoning diversity, stabilizes training, and avoids the observed late-stage mode collapse.
\end{itemize}

\section{Preliminaries and Related Work}

\subsection{RL-Based Post-Training for LLMs}

Post-training has been widely used to align large language models (LLMs) with human preferences. Representative approaches include reinforcement learning from human feedback (RLHF)~\citep{lee2024rlaif,ouyang2022training} and direct preference optimization (DPO)~\citep{rafailov2023direct}, which optimizes a preference objective without explicit reinforcement learning. More recently, reinforcement learning with verifiable rewards (RLVR) has become an effective paradigm for improving LLM reasoning~\citep{guo2025deepseek,shao2024deepseekmath,team2025kimi,dai2025s,meng2025mm,zhou2026one}. Group Relative Policy Optimization (GRPO) is a representative RLVR algorithm that estimates the advantage of each rollout relative to other responses sampled for the same problem, avoiding a separately trained critic~\citep{shao2024deepseekmath,guo2025deepseek}. Subsequent studies improve GRPO-style training through modifications to the objective, advantage normalization, clipping, and sampling~\citep{yu2025dapo,liu2025understanding,yue2025vapo,zheng2025group,chu2025gpg,zhao2026reinforced,zhou2026look}.

Given a problem $x$, the policy samples a group of on-policy rollouts $\{y_i\}_{i=1}^{G}$ with rewards $\{r_i\}_{i=1}^{G}$. GRPO assigns each rollout the group-relative advantage
\begin{equation}
    A_i = \frac{r_i-\mu_G}{\sigma_G},
\end{equation}
where $\mu_G$ and $\sigma_G$ are the mean and standard deviation of group rewards. Although the policy objective is optimized over individual tokens, every token in rollout $y_i$ shares the same response-level advantage $A_i$. GRPO therefore provides no explicit token-level credit assignment for distinguishing valid reasoning steps from erroneous ones. In particular, an incorrect rollout may receive a negative advantage over its entire trajectory even when its prefix contains valid reasoning.

\subsection{On-Policy Self-Distillation}

Classical knowledge distillation trains a student model to imitate outputs generated by a stronger teacher~\citep{hinton2015distilling,gu2024minillm,kim2016sequence}. For LLM reasoning, such supervision is typically off-policy because the targets are generated by the teacher rather than sampled from the student's current policy. In contrast, on-policy self-distillation samples rollouts from the current student and uses a self-teacher conditioned on additional information to provide token-level supervision along these on-policy trajectories~\citep{agarwal2024policy,lu2025onpolicydistillation}. OPSD~\citep{zhao2026self} and SDFT~\citep{shenfeld2026self} condition the self-teacher on reference solutions, while SDPO~\citep{hubotter2026reinforcement} uses environmental feedback or successful rollouts to guide unsuccessful trajectories.

At token position $t$, let
\begin{equation}
    p_{i,t}^{\mathrm{S}}
    =
    \pi_{\theta}(\cdot\mid x,y_{i,<t}),
    \qquad
    p_{i,t}^{\mathrm{T}}
    =
    \pi_{\theta}(\cdot\mid x,c_i,y_{i,<t}),
\end{equation}
denote the student and self-teacher distributions, respectively, where $c_i$ is the additional context. SDPO minimizes a token-level distillation objective over the entire rollout:
\begin{equation}
\label{eq:sdpo_loss}
    \mathcal{L}_{\mathrm{SDPO}}(\theta)
    =
    \sum_{i=1}^{G}\sum_{t=1}^{|y_i|}
    D\left(
        p_{i,t}^{\mathrm{S}},
        \operatorname{stopgrad}\!\left[p_{i,t}^{\mathrm{T}}\right]
    \right),
\end{equation}
where $D$ may be KL divergence, reverse KL divergence, or Jensen--Shannon divergence. Across existing methods, the self-teacher may share parameters with the student, remain fixed, or be updated as an exponential moving average~\citep{li2026rethinking}.

Compared with GRPO, SDPO provides a teacher-derived distribution at every token position rather than sharing a response-level advantage across the rollout. However, standard OPSD applies this dense supervision to the entire response. Conditioning the self-teacher on a complete correct solution may encourage direct trajectory imitation, while full-response distillation may impose unnecessary gradients on already-valid prefixes, suppressing reasoning diversity and destabilizing training.

\MethodName addresses both limitations through reflection-guided, error-localized distillation. It extracts concise, rollout-specific guidance---a corrective idea for an incorrect rollout or a key successful insight for a correct one---rather than directly exposing the self-teacher to a complete correct trajectory. For an incorrect rollout, it identifies the sentence containing the first reasoning error, masks the preceding valid prefix, and applies distillation only to the suffix beginning with that sentence. Thus, \MethodName retains the dense supervision of OPSD while explicitly localizing where correction should begin.

\begin{figure*}[t]
\centering
\includegraphics[width=1\linewidth]{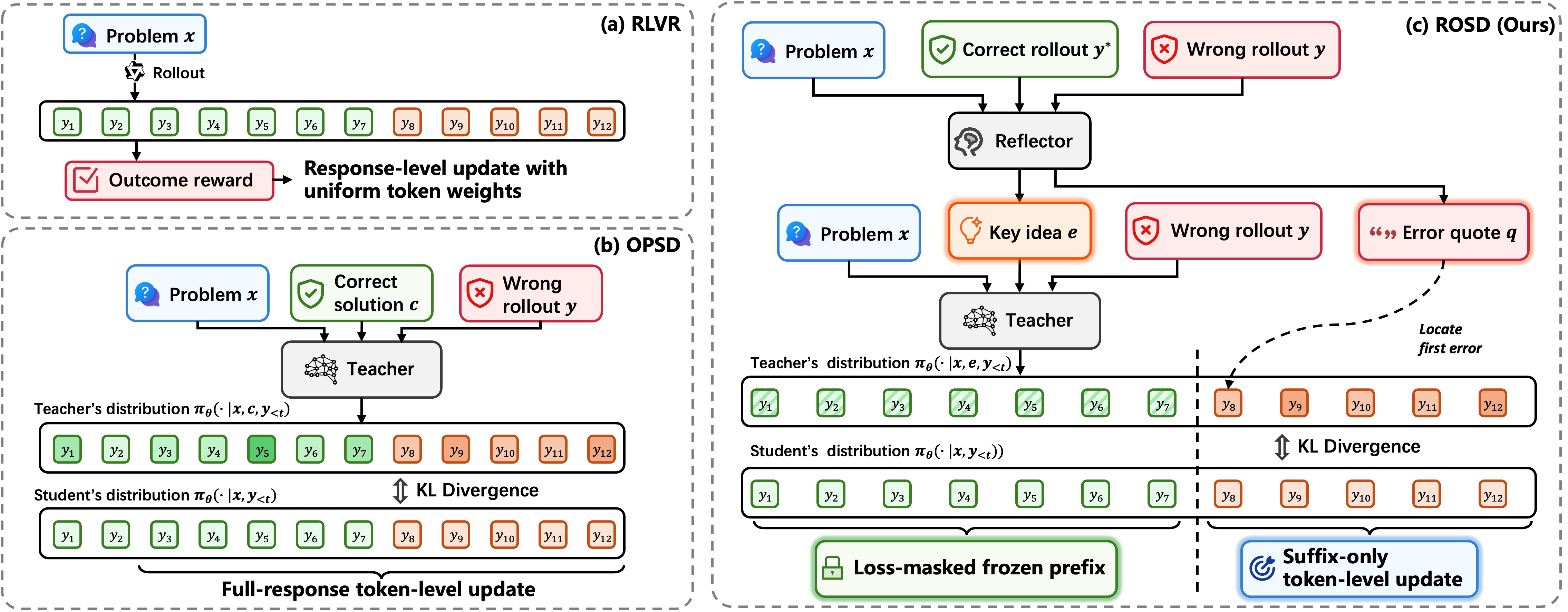}
\caption{
Conceptual comparison of the learning signals in RLVR, standard OPSD, and \MethodName, illustrated using an erroneous rollout.
\textbf{(a)} RLVR provides only an outcome-level reward for the complete rollout.
\textbf{(b)} Standard OPSD conditions the self-teacher on a complete correct solution and distills the entire response, including its already-valid prefix.
\textbf{(c)} \MethodName uses a self-reflector to derive a corrective idea and identify the sentence containing the first reasoning error. The corrective idea guides the self-teacher, while the diagnosed error boundary masks out the valid prefix and restricts token-level distillation to the suffix beginning with the first erroneous sentence.
Green and orange tokens denote the valid prefix and error-affected suffix, respectively; hatched tokens indicate positions masked out from the distillation loss.
}
\label{fig:method_overview}
\vspace{-0.3cm}
\end{figure*}

\section{Proposed Method: \MethodName}\label{sec:method}


\MethodName shifts on-policy self-distillation from rigid trajectory imitation to distilling transferable reasoning insights for targeted reasoning correction.
As illustrated in Figure~\ref{fig:method_overview}, it differs from both the outcome-level supervision of RLVR and the full-response distillation of standard OPSD.
Given an input reasoning task, the policy first generates candidate on-policy rollouts.
For each erroneous rollout, a self-reflector derives a corrective idea from a verified solution and identifies the first incorrect reasoning step (Section~\ref{sec:diagnosis}).
The corrective idea guides the self-teacher toward targeted supervision, while the diagnosed error boundary enables \MethodName to mask out the distillation loss over the valid prefix and apply token-level distillation only from the first error onward (Section~\ref{sec:localized_distillation}).
\MethodName uses the same base model for student model,  reflection and self-teaching, requiring no auxiliary models or additional trainable parameters.

\subsection{Outcome-Aware Self-Reflection}\label{sec:diagnosis}

In standard SDPO, the auxiliary context $c$ can be a complete solution. Although this provides the self-teacher with strong information about a correct solution, it does not reveal where or why the student's on-policy rollout fails. Consequently, the self-teacher can be overly shaped by the reference trajectory, including its reasoning path, format, and stylistic details. In contrast, \MethodName does not directly expose the self-teacher to the complete reference solution. Instead, it introduces an outcome-aware self-reflector that extracts rollout-specific reasoning insights: it diagnoses how an incorrect rollout should be corrected while summarizing the key idea underlying a correct rollout.

For each problem $x$, the current policy samples $G$ on-policy rollouts:
\begin{equation}
    \mathcal{Y}(x)=\{y_1,\ldots,y_G\}.
\end{equation}
Based on final-answer correctness, we partition them into correct rollouts $\mathcal{Y}^{+}(x)$ and incorrect rollouts $\mathcal{Y}^{-}(x)$. Consistent with group-based reinforcement learning, we retain only mixed-outcome groups containing both correct and incorrect rollouts~\citep{shao2024deepseekmath,yu2025dapo}. \MethodName then constructs different self-reflection prompts for incorrect and correct rollouts.

\paragraph{Contrastive reflection for incorrect rollouts.}
For each incorrect rollout $y^{-}\in\mathcal{Y}^{-}(x)$, we select the shortest correct rollout from the same group:
\begin{equation}
    y^{*}=\arg\min_{y\in\mathcal{Y}^{+}(x)} |y|,
\end{equation}
and use it as a contrastive basis for diagnosing where the incorrect reasoning deviates. Selecting the shortest correct rollout encourages the reflection to identify a concise and efficient corrective direction. Given the problem $x$, the correct rollout $y^{*}$, and the incorrect rollout $y^{-}$, the self-reflector produces a corrective idea $e^{-}$ and an error quote $q$:
\begin{equation}\label{eq:wrong_reflection}
    (e^{-},q) \sim \pi_{\theta}(\cdot \mid x,y^{*},y^{-}).
\end{equation}
The corrective idea $e^{-}$ explains why the rollout fails and how its reasoning can be repaired. The error quote $q$ is the exact sentence from $y^{-}$ that contains the first identified reasoning error.  It serves only as a localization anchor and is later aligned with the student rollout to determine where localized distillation begins.

\paragraph{Positive reflection for correct rollouts.}
For each correct rollout $y^{+}\in\mathcal{Y}^{+}(x)$, no error localization is required. Instead of directly using the complete rollout as teacher context, the self-reflector summarizes the key idea underlying its successful reasoning:
\begin{equation}\label{eq:correct_reflection}
    e^{+} \sim \pi_{\theta}(\cdot \mid x,y^{+}).
\end{equation}
This positive reflection retains the transferable reasoning insight from a successful rollout without encouraging direct imitation of its complete trajectory.

\paragraph{Key insight.}
The extracted insight, either $e^{-}$ or $e^{+}$, is subsequently used as the auxiliary context for the self-teacher. Self-reflection acts as an information bottleneck that converts complete solutions into concise, rollout-specific reasoning guidance. For incorrect rollouts, it further separates \emph{what to correct}, represented by $e^{-}$, from \emph{where to begin correction}, identified by $q$. This separation enables targeted teacher guidance while preserving reasoning that is already valid.

The self-reflector shares the same model weights as the policy and self-teacher, requiring no auxiliary model or additional trainable parameters. The complete self-reflection prompt templates are provided in Appendix~\ref{app:reflection_prompt_templates} (Figure~\ref{fig:reflection_prompt}).

\subsection{Error-Localized Self-Distillation}\label{sec:localized_distillation}

Even with rollout-specific reflection, applying self-distillation to the entire response remains unnecessary. Because the self-teacher is conditioned on additional reasoning guidance, its distribution may differ from the student's natural generation even over valid reasoning steps. Full-response distillation may therefore penalize valid alternative prefixes, suppress reasoning diversity, and increase the risk of mode collapse. For an incorrect rollout, we instead use the error quote $q$ to preserve the valid prefix and restrict distillation to the suffix beginning with the first erroneous sentence.

Specifically, for each incorrect rollout $y^{-}\in\mathcal{Y}^{-}(x)$, \MethodName matches the error quote $q$ against the original rollout and identifies the token index at which the quoted sentence begins:
\begin{equation}
    k=\mathrm{Locate}(q,y^{-}).
\end{equation}
If the quote cannot be matched, we fall back to full-response distillation by setting $k=1$. We then define the token-level mask
\begin{equation}
m_t =
\begin{cases}
0, & t < k,\\
1, & t \ge k.
\end{cases}
\end{equation}
Thus, tokens in the valid prefix do not contribute to the distillation loss, whereas all tokens from the first erroneous sentence onward are retained. For a correct rollout, no error localization is needed, and we set $m_t=1$ for all response tokens.

Let $e$ denote the reflection associated with the rollout, i.e., $e=e^{-}$ for an incorrect rollout and $e=e^{+}$ for a correct rollout. The reflection conditions the self-teacher on targeted reasoning guidance. At each response position $t$, the student and self-teacher distributions are respectively
\begin{equation}
    p_t^{\mathrm{S}}
    = \pi_{\theta}(\cdot \mid x,y_{<t}),
    \qquad
    p_t^{\mathrm{T}}
    = \pi_{\theta}(\cdot \mid x,e,y_{<t}).
\end{equation}
Both distributions are evaluated over the same rollout prefix $y_{<t}$, while gradients through the self-teacher are stopped. Following prior work~\citep{hubotter2026reinforcement,li2026unifying}, we instantiate the token-level divergence using Jensen--Shannon divergence (JSD):
\begin{equation}
\mathcal{L}_{\mathrm{\MethodName}}(\theta)
=
\sum_{t=1}^{T}
m_t\,
\mathrm{JSD}\left(
p_t^{\mathrm{S}}
\,\middle\|\,
\operatorname{stopgrad}\!\left[p_t^{\mathrm{T}}\right]
\right).
\end{equation}
The mask $m_t$ ensures that, for an incorrect rollout, distillation begins only at the first diagnosed error, thereby preserving the valid reasoning prefix while correcting the subsequent reasoning trajectory. The student and self-teacher prompt templates are provided in Appendix~\ref{app:prompt_templates}.

\section{Experiments}\label{sec:experiments}

\subsection{Experimental Setup}\label{sec:experimental_setup}

\paragraph{Models.}
We evaluate our method across two diverse and representative model families: Qwen3-4B, Qwen3-8B~\citep{yang2025qwen3}, and Llama-3.2-3B-Instruct~\citep{grattafiori2024llama}.
For \MethodName and OPSD baselines,
the student,
self-teacher,
and self-reflector share the same base model.
Thus,
the performance gains come from training with a self-teacher conditioned on privileged information,
rather than from using a larger external teacher.

\paragraph{Datasets.}
Following prior work~\citep{hubotter2026reinforcement}, we train models separately on five datasets.
The first four datasets are science question-answering benchmarks from the reasoning subset (L3) of SciKnowEval~\citep{feng2024sciknoweval},
covering undergraduate-level scientific reasoning in Chemistry, Physics, Biology, and Materials Science.
In addition, we include a tool-use dataset from ToolAlpaca~\citep{tang2023toolalpaca},
where the model is required to map a tool-API specification and a user request to the correct tool call.
Each of the five datasets is split into a training set and a test set.
Models are trained on each domain and evaluated on its corresponding test set.
The statistics of each dataset are shown in Appendix~\ref{app:dataset_statistics}.

\paragraph{Baselines.}
We compare \MethodName with two baselines in the standard RLVR setting, where no human-labeled golden solutions are available and models are trained only with feedback from a verifier.
(1) For the GRPO baseline, we use a strengthened implementation of GRPO following recent best practices~\citep{khatri2025art}, representing outcome-level reinforcement learning.
This implementation incorporates recent refinements to standard GRPO, including asymmetric clipping~\citep{yu2026dapo}, unbiased normalization~\citep{liu2025understanding}, and off-policy correction for efficient inference-based rollout generation~\citep{yao2025your}.
(2) For the on-policy self-distillation baseline, we use SDPO~\citep{hubotter2026reinforcement}, adopting its official implementation and hyperparameter configuration for the verifier-only RLVR setting.
SDPO uses successful rollouts sampled by the current policy as additional self-teacher context, thereby providing denser token-level guidance.
We extend training to the epoch budgets specified in Appendix~\ref{app:implementation_details} to examine longer-term training dynamics.
We do not include other OPSD methods, such as OPSD~\citep{zhao2026self} and SDFT~\citep{shenfeld2026self}, in the main comparison because they rely on high-quality human-labeled golden solutions.

\paragraph{Implementation details.}
All methods are trained with the same data splits,
answer verifiers,
and evaluation prompts.
For each prompt,
we sample 8 on-policy rollouts during training.
Following prior work~\citep{hubotter2026reinforcement, li2026rethinking}, evaluation uses 16 sampled responses per problem and reports the mean accuracy,
denoted as mean@16.
Additional training hyperparameters and distillation settings are provided in Appendix~\ref{app:implementation_details}.

\subsection{Main Results}\label{sec:main_results}

\begin{table*}[t]
\centering
\small
\setlength{\tabcolsep}{4pt}
\begin{tabular}{@{}ll*{6}{l}@{}}
\toprule
Model & Method & Material & Physics & Biology & Chemistry & ToolUse & Average \\
\midrule
\multirow{4}{*}{Qwen3-4B}
& Base & 62.90 & 60.63 & 31.63 & 42.92 & 58.55 & 51.32 \\
& GRPO & \textbf{81.45} & \underline{74.61} & 53.75 & \underline{77.53} & \underline{61.95} & \underline{69.86} \\
& SDPO & 73.60 & 68.52 & \underline{55.38} & 76.93 & 60.66 & 67.02 \\
& \MethodName & \underline{80.18} & \textbf{76.56} & \textbf{57.50} & \textbf{82.47} & \textbf{67.46} & \textbf{72.83} \\
\midrule
\multirow{4}{*}{Qwen3-8B}
& Base & 59.71 & 59.06 & 32.75 & 41.28 & 57.72 & 50.10 \\
& GRPO & 77.46 & \underline{76.09} & 56.88 & \underline{80.77} & \underline{68.75} & 71.99 \\
& SDPO & \underline{78.46} & \textbf{76.41} & \textbf{61.13} & 80.71 & 65.81 & \underline{72.50} \\
& \MethodName & \textbf{79.52} & 75.16 & \underline{60.38} & \textbf{83.07} & \textbf{69.12} & \textbf{73.45} \\
\midrule
\multirow{4}{*}{\shortstack{Llama-3.2-\\3B-Instruct}}
& Base & 0.01 & 0.01 & 0.00 & 0.01 & 0.00 & 0.01 \\
& GRPO & 46.14 & \textbf{48.04} & \underline{45.37} & \underline{65.56} & \textbf{58.36} & \underline{52.69} \\
& SDPO & \underline{45.47} & 46.01 & 42.25 & 61.78 & 55.97 & 50.30 \\
& \MethodName & \textbf{46.76} & \underline{47.57} & \textbf{49.12} & \textbf{66.36} & \underline{57.07} & \textbf{53.38} \\
\bottomrule
\end{tabular}
\caption{Reasoning performance across scientific and tool-use benchmarks on Qwen3-4B, Qwen3-8B, and Llama-3.2-3B-Instruct.
``Base'' denotes the untuned base model.
All methods are trained to convergence, and scores are reported as the mean@16(\%).
\textbf{Boldface} indicates the best result,
and \underline{underline} indicates the second-best result.}
\label{tab:in_domain_results}
\vspace{-10pt}
\end{table*}

Table~\ref{tab:in_domain_results} reports the results across five tasks using Qwen3-4B, Qwen3-8B, and Llama-3.2-3B-Instruct. We highlight two main findings.

\paragraph{\MethodName performs consistently across model architectures.}
\MethodName achieves the best average performance on all three backbones and consistently outperforms both GRPO and SDPO across model families and scales. In contrast, the relative performance of SDPO depends more strongly on model capacity: it slightly outperforms GRPO on Qwen3-8B but falls behind GRPO on the smaller Qwen3-4B and Llama-3.2-3B models. These results suggest that dense token-level supervision alone does not guarantee improvements over outcome-level RL, particularly when model capacity is limited. \MethodName instead delivers more consistent gains across the evaluated backbones.

\paragraph{\MethodName performs well across diverse task domains.}
The two self-distillation methods also differ substantially across individual tasks. \MethodName achieves or approaches the best performance on both scientific question answering and tool use, whereas SDPO exhibits pronounced degradation in several settings, including Physics and Materials Science on Qwen3-4B. This discrepancy is consistent with the limitations of full-response distillation, which may impose unnecessary supervision on already-valid reasoning prefixes. By extracting rollout-specific guidance and applying distillation only to the suffix beginning with the first erroneous sentence, \MethodName preserves valid prefixes while correcting error-affected continuations.

\subsection{Training Dynamics and Stability}\label{sec:training_dynamics}

\begin{figure}[t]
\centering
\includegraphics[width=\textwidth]{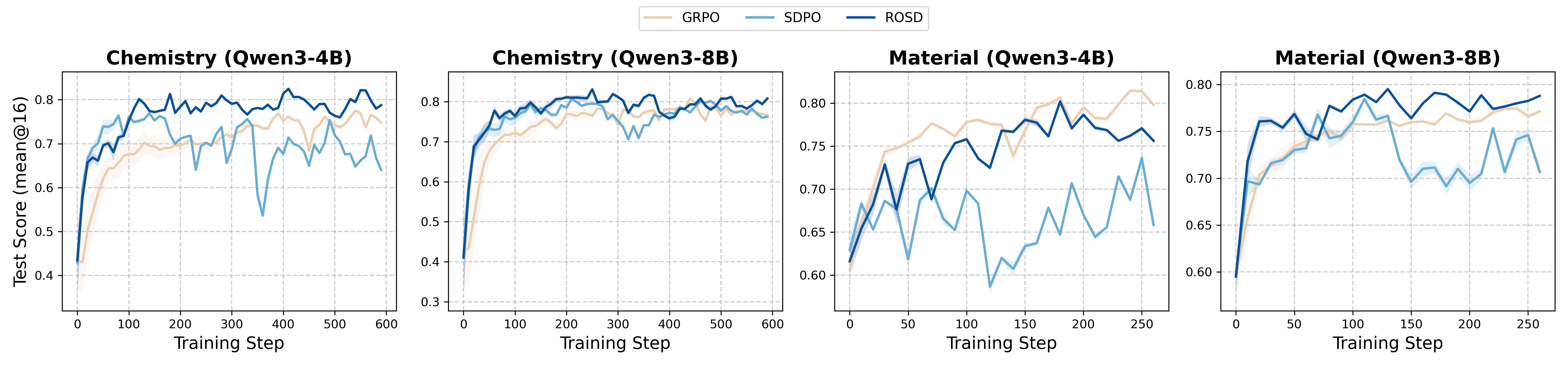}
\caption{Test-set performance during training on Chemistry and Materials Science using Qwen3-4B and Qwen3-8B. We report mean@16, with shaded regions indicating standard deviation across sampled responses.}
\label{fig:training_dynamics_test_score}
\includegraphics[width=\textwidth]{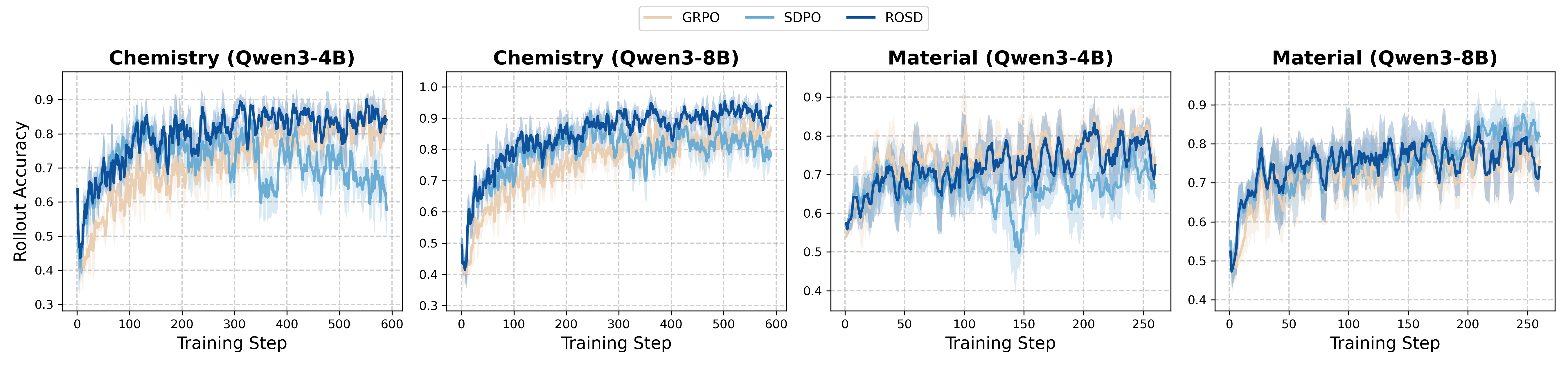}
\caption{On-policy rollout accuracy during training on Chemistry and Materials Science using Qwen3-4B and Qwen3-8B. Curves show rolling averages over a five-step window, with shaded regions indicating standard deviation.}
\label{fig:training_dynamics_rollout_accuracy}
\vspace{-10pt}
\end{figure}

To examine the optimization process, we analyze Qwen3-4B and Qwen3-8B on Chemistry and Materials Science. We report test-set mean@16 in Figure~\ref{fig:training_dynamics_test_score} and on-policy rollout accuracy in Figure~\ref{fig:training_dynamics_rollout_accuracy}. The results reveal two main patterns.

\paragraph{Dense self-distillation accelerates early learning.}
During early training, both SDPO and \MethodName improve rollout accuracy considerably faster than GRPO. GRPO assigns each rollout a shared advantage derived from its outcome, providing no explicit token-level credit assignment. In contrast, self-distillation provides teacher-derived supervision at every token position, enabling faster initial optimization.

\paragraph{\MethodName remains stable during extended training.}
Despite its rapid initial improvement, SDPO exhibits substantial degradation later in training. On Qwen3-4B, its rollout accuracy and test performance collapse on both Chemistry and Materials Science. On Materials Science with Qwen3-8B, its rollout accuracy remains high while its test score steadily declines, indicating a widening generalization gap consistent with overfitting. In contrast, \MethodName maintains stable or improving performance across the evaluated settings. These results suggest that masking valid prefixes and restricting distillation to the error-affected suffix mitigates the instability associated with full-response distillation.

\subsection{Response Diversity Analysis}\label{sec:diversity_analysis}

\begin{table}[t]
\centering
\small
\setlength{\tabcolsep}{3pt}
\resizebox{0.95\textwidth}{!}{%
\begin{tabular}{l | cc | cc | cc | cc | cc | cc}
\toprule
\multirow{2}{*}{\textbf{Method}} & \multicolumn{2}{c|}{\textbf{Chemistry}} & \multicolumn{2}{c|}{\textbf{Material}} & \multicolumn{2}{c|}{\textbf{Physics}} & \multicolumn{2}{c|}{\textbf{Biology}} & \multicolumn{2}{c|}{\textbf{ToolUse}} & \multicolumn{2}{c}{\textbf{Average}} \\
& SB-4 $\downarrow$ & Dist-2 $\uparrow$ & SB-4 $\downarrow$ & Dist-2 $\uparrow$ & SB-4 $\downarrow$ & Dist-2 $\uparrow$ & SB-4 $\downarrow$ & Dist-2 $\uparrow$ & SB-4 $\downarrow$ & Dist-2 $\uparrow$ & SB-4 $\downarrow$ & Dist-2 $\uparrow$ \\
\midrule
Base & 80.01 & 20.24 & 95.09 & 8.75 & 90.19 & 10.08 & 80.79 & 19.99 & 97.28 & 9.37 & 88.67 & 13.69 \\
GRPO & \textbf{87.69} & \textbf{15.02} & \underline{90.97} & \underline{12.90} & \textbf{90.68} & \underline{11.89} & \textbf{93.70} & \textbf{8.82} & 95.37 & 11.52 & \textbf{91.68} & \underline{12.03} \\
SDPO & 92.99 & 12.30 & \textbf{89.55} & 12.69 & 94.13 & 11.00 & 99.20 & 6.13 & \underline{92.56} & \underline{12.65} & 93.69 & 10.95 \\
\MethodName (Ours) & \underline{90.87} & \underline{14.69} & 91.17 & \textbf{15.10} & \underline{91.82} & \textbf{13.78} & \underline{98.61} & \underline{6.84} & \textbf{90.48} & \textbf{14.14} & \underline{92.59} & \textbf{12.91} \\
\bottomrule
\end{tabular}%
}
\caption{Reasoning diversity analysis on Qwen3-4B at training step 140 across five domains.
We report Self-BLEU-4 (SB-4, \%, lower indicates higher diversity) and Distinct-2 (Dist-2, \%, higher indicates higher lexical richness) on generated rollouts.
\textbf{Boldface} indicates the best result, and \underline{underline} indicates the second-best result among post-training methods (excluding Base).}
\label{tab:diversity_results}
\vspace{-5pt}
\end{table}

We further examine whether full-response self-distillation reduces response diversity. Specifically, we evaluate rollouts generated by Qwen3-4B at the final training checkpoint across all five tasks. Following prior work, we use two complementary metrics: \textbf{Self-BLEU-4}~\citep{zhu2018texygen}, which measures similarity among responses generated for the same prompt (lower is more diverse), and \textbf{Distinct-2}~\citep{li2016diversity}, which measures the proportion of unique bigrams among all generated bigrams (higher is more diverse).

\paragraph{\MethodName better preserves response diversity.}
As shown in Table~\ref{tab:diversity_results}, SDPO generally produces higher Self-BLEU-4 and lower Distinct-2 than GRPO, with particularly low diversity on Biology. This pattern indicates that full-response distillation can make generated trajectories more homogeneous. In contrast, \MethodName achieves higher Distinct-2 across all five tasks and generally lower Self-BLEU-4 than SDPO. These results suggest that masking valid prefixes reduces unnecessary imitation and better preserves variation among generated responses.

\subsection{Ablation Study}\label{sec:ablation_study}

\begin{table}[t]
\centering
\scriptsize
\setlength{\tabcolsep}{4pt}
\renewcommand{\arraystretch}{1}
\begin{tabular}{llcccccc}
\toprule
\textbf{Model} & \textbf{Method} & \textbf{Material} & \textbf{Physics} & \textbf{Biology} & \textbf{Chemistry} & \textbf{ToolUse} & \textbf{Average} \\
\midrule
\multirow{4}{*}{Qwen3-4B}
& SDPO & 73.60 & 68.52 & \underline{55.38} & 76.93 & 60.66 & 67.02 \\
& \MethodName & \textbf{80.18} & \textbf{76.56} & \textbf{57.50} & \textbf{82.47} & \textbf{67.46} & \textbf{72.83} \\
& \quad $\vdash$ w/o Refl. & \underline{75.27} & 73.13 & 51.13 & 78.84 & 59.47 & 67.56 \\
& \quad $\vdash$ w/o LocDist. & 75.13 & \underline{75.39} & 54.50 & \underline{80.12} & \underline{63.51} & \underline{69.73} \\
\midrule
\multirow{4}{*}{Qwen3-8B}
& SDPO & 78.46 & \textbf{76.41} & \textbf{61.13} & 80.71 & 65.81 & \underline{72.50} \\
& \MethodName & \underline{79.52} & 75.16 & \underline{60.38} & \underline{83.07} & \textbf{69.12} & \textbf{73.45} \\
& \quad $\vdash$ w/o Refl. & 73.80 & 71.41 & 59.13 & \textbf{83.10} & 65.07 & 70.50 \\
& \quad $\vdash$ w/o LocDist. & \textbf{79.99} & \underline{76.25} & 56.50 & 81.76 & \underline{67.19} & 72.34 \\
\midrule
\multirow{4}{*}{\shortstack{Llama-3.2-\\3B-Instruct}}
& SDPO & 45.47 & 46.01 & 42.25 & 61.78 & 55.97 & 50.30 \\
& \MethodName & \textbf{46.76} & \textbf{47.57} & \textbf{49.12} & \textbf{66.36} & \textbf{57.07} & \textbf{53.38} \\
& \quad $\vdash$ w/o Refl. & 45.23 & \underline{46.54} & 45.65 & 63.24 & 54.47 & 51.03 \\
& \quad $\vdash$ w/o LocDist. & \underline{46.02} & 45.71 & \underline{47.84} & \underline{65.92} & \underline{56.31} & \underline{52.26} \\
\bottomrule
\end{tabular}
\caption{``w/o Refl.'' denotes without reflection guidance; ``w/o LocDist.'' denotes without localized distillation.
\textbf{Boldface} indicates the best result, and \underline{underline} indicates second-best.}
\label{tab:ablation_results}
\vspace{-10pt}
\end{table}




To dissect the contribution of each core component in \MethodName, we evaluate two ablation variants across all five reasoning benchmarks on the three backbones (Table~\ref{tab:ablation_results}):
(1) \textbf{w/o Reflection}, which retains error localization to identify the logical inflection point but conditions the self-teacher directly on the raw reference solution rather than generating targeted self-reflective diagnosis;
and (2) \textbf{w/o Localized Distillation}, which preserves reflective guidance but applies distillation loss indiscriminately across the entire response trajectory, omitting prefix-preserving selective supervision.

As shown in Table~\ref{tab:ablation_results}, removing reflective guidance leads to a substantial performance drop across all three backbones, with average accuracy falling close to that of standard SDPO.
This indicates that raw reference imitation provides diffuse and uninformative supervision, whereas targeted self-reflection explicitly diagnoses logical flaws to deliver sharp, semantically focused corrective signals.
Furthermore, even with reflective teacher guidance, omitting localized distillation consistently underperforms full \MethodName across all architectures.
Forcing the model to distill already valid prefixes introduces redundant imitation penalties that stifle exploration and induce mode collapse, whereas confining supervision strictly to the erroneous suffix leaves sound prefixes unperturbed, stabilizing optimization and maximizing reasoning gains.
Together, these findings demonstrate that reflective diagnosis and localized prefix preservation are complementary and jointly drive the superior performance of \MethodName.



\section{Discussion and Conclusion}\label{sec:conclusion}

\paragraph{Discussion.}
Our results reveal a central trade-off in reasoning post-training: outcome-level RLVR provides coarse feedback, whereas standard on-policy self-distillation, as represented by SDPO, offers dense supervision but may unnecessarily alter valid reasoning prefixes. The effectiveness of \MethodName, however, depends on the quality of self-reflection and error localization, which may be less reliable for ambiguous or highly specialized problems. Moreover, sentence-level localization is a practical approximation to exact token-level error attribution. We provide further analyses of cross-domain generalization (Appendix~\ref{app:cross_domain_results}), training efficiency (Appendix~\ref{app:training_efficiency}), response length dynamics (Appendix~\ref{app:response_length}), error localization dynamics (Appendix~\ref{app:error_localization_dynamics}), and qualitative examples (Appendix~\ref{app:case_study_biology}).

\paragraph{Conclusion.}
We introduced \MethodName, a reflective on-policy self-distillation framework for improving LLM reasoning. \MethodName extracts rollout-specific guidance through self-reflection and applies distillation only to the suffix beginning with the first erroneous sentence, thereby correcting error-affected continuations while preserving valid prefixes. Across five tasks, \MethodName achieves the best overall average performance across the three backbones, outperforming both GRPO and SDPO. Compared with SDPO, \MethodName better preserves response diversity and improves training stability. Additional results demonstrate stronger cross-domain generalization than SDPO. Overall, our findings highlight the importance of combining dense token-level guidance with error-aware localization for reliable and generalizable post-training.

\subsection*{AI use statement}

In this work, we used generative AI tools for implementing methods and assisting with translation.
We also used an LLM as a judge to evaluate the model rollouts, as detailed in Appendix \ref{app:token_kl_analysis}.
We have not used generative AI tools for proposing or refining hypotheses, designing or providing feedback on research methodology or experiments, cleaning and reformatting datasets, or interpreting results.
Generating synthetic data sets, helping develop theoretical models or conceptual frameworks, formulating mathematical claims, providing critical ingredients for proving mathematical claims, and assisting in the writing of proofs are not applicable to this work.
Additionally, we used generative AI tools for finding related work.
We have reviewed all AI-assisted work: all AI-assisted code implementations were manually inspected, debugged, and verified through unit and integration testing by the authors, and all suggested citations were independently read and confirmed.
We take responsibility for the final content of this work, including text, claims or artifacts produced with the aid of generative AI.
\section*{Ethics Statement}

This work studies post-training methods for improving the reasoning capabilities of large language models. It does not involve human participants, crowdsourcing, or the collection of personal or sensitive information. All experiments use publicly available benchmark datasets, which are used in accordance with their licenses and intended research purposes. As with other methods that improve reasoning and tool-use capabilities, \MethodName may strengthen both beneficial and potentially harmful downstream capabilities. Our experiments are limited to benchmark reasoning and tool-use tasks and should not be interpreted as establishing the safety, fairness, or reliability of the resulting models in real-world deployment. The trained models may also inherit biases and other limitations from their pretrained backbones and benchmark data. Deployment should therefore be accompanied by application-specific safety evaluation, appropriate access controls, and continued monitoring. We adhere to the ICLR Code of Ethics and acknowledge these potential limitations and risks.
\section*{Reproducibility Statement}

We provide the complete formulation and training procedure of \MethodName in Section~\ref{sec:method}. Section~\ref{sec:experimental_setup} describes the model backbones, datasets, baselines, evaluation metrics, and the shared training and evaluation protocol used for all compared methods. Appendix~\ref{app:dataset_statistics} reports the dataset composition and train--test splits, while Appendix~\ref{app:implementation_details} specifies the computing infrastructure, training budgets, batch size, sequence-length limits, evaluation frequency, and distillation hyperparameters. The self-reflection and teacher prompt templates are provided in Appendix~\ref{app:reflection_prompt_templates}. An anonymous implementation is available at \url{https://anonymous.4open.science/r/ROSD}. 

\bibliography{custom}
\bibliographystyle{iclr2027_conference}

\newpage
\appendix

\section{Dataset Statistics}
\label{app:dataset_statistics}

Table~\ref{tab:dataset_setup} reports the number of training and held-out examples used in our experiments.
The science question-answering and ToolUse domains are used for training and evaluation.
The AIME2024 dataset is used only as an additional held-out reasoning benchmark and is not used for training.

\begin{table}[h]
\centering
\small

\begin{tabular}{lrr}
\toprule
Datasets & Train & Test \\
\midrule
Biology & 450 & 50 \\
Chemistry & 1,890 & 210 \\
Material & 841 & 94 \\
Physics & 720 & 80 \\
ToolUse & 4,046 & 68 \\
AIME2024 & - & 30 \\
\bottomrule
\end{tabular}
\caption{Dataset statistics used in our experiments.}
\label{tab:dataset_setup}
\end{table}

\section{Implementation Details}
\label{app:implementation_details}

\paragraph{Infrastructure.} All experiments are implemented on top of the \texttt{verl} library~\citep{sheng2025hybridflow}.
We use the distributed actor-rollout training pipeline provided by \texttt{verl},
with FSDP actor training and vLLM-based rollout generation.
The experiments are run on 8 NVIDIA A800 80G GPUs.

\paragraph{Training.}
Unlike prior methods~\citep{hubotter2026reinforcement} that train under a fixed time budget,
we use an epoch budget large enough for the methods to converge in our setting.
We train for 10 epochs on the science question-answering tasks and 5 epochs on the ToolUse task,
using a training batch size of 32.
The maximum length of the reflection prompt is 8k tokens,
and the maximum length of the generated reflection is 4k tokens.
All algorithms are evaluated every 10 training steps.
Both SDPO and \MethodName use Jensen--Shannon divergence with $\alpha=0.5$ and top-$k$ distillation with $k=100$.
Following the original implementation and hyperparameter configuration, SDPO updates its self-teacher via an exponential moving average (EMA) with an update rate of $0.05$ at every training step.
In \MethodName, the self-teacher and self-reflector remain frozen throughout each training run.

\section{Self-Reflection Prompt Templates}
\label{app:reflection_prompt_templates}

Figure~\ref{fig:reflection_prompt} presents the self-reflection prompt templates used by \MethodName.
The system prompt provides task-level instructions to the reflector.
For wrong rollouts, the reflector receives the problem, the shortest correct rollout, and the wrong rollout to identify the first erroneous span and provide a corrective explanation.
For correct rollouts, the reflector summarizes why the reasoning path is valid and what key idea leads to the correct answer.

\begin{figure*}[t]
\centering
\footnotesize
\fcolorbox{black}{gray!4}{%
\begin{minipage}{\dimexpr\linewidth-2\fboxsep-2\fboxrule\relax}
\textbf{System prompt for both rollouts.}\\
You are a careful tutor.
Respond strictly in the required format and nothing else.

\medskip
\textbf{User prompt for wrong rollouts.}\\
\texttt{[Problem]}\\
\textcolor{blue}{\{problem\}}\\
\texttt{[Correct Solution]}\\
\textcolor{blue}{\{correct\_rollout\}}\\
\texttt{[Incorrect Solution]}\\
\textcolor{blue}{\{wrong\_rollout\}}\\
The correct final answer is \textcolor{blue}{\{answer\}}.
Diagnose the [Incorrect Solution] by responding strictly in the following format:\\
\texttt{\textless error\_quote\textgreater}\\
(Extract the EXACT substring from the [Incorrect Solution] where the reasoning first goes wrong)\\
\texttt{\textless/error\_quote\textgreater}\\
\texttt{\textless explanation\textgreater}\\
(Explain the exact mistake in the quote, how to fix it, and what the correct logic is)\\
\texttt{\textless/explanation\textgreater}

\medskip
\textbf{User prompt for correct rollouts.}\\
\texttt{[Problem]}\\
 \textcolor{blue}{\{problem\}}\\
\texttt{[Correct Solution]}\\
\textcolor{blue}{\{correct\_rollout\}}\\
Explain why this reasoning is valid and why the final answer should be \textcolor{blue}{\{answer\}}. Respond strictly in the following format:\\
\texttt{\textless explanation\textgreater}\\
(Your explanation of why the logic is correct)\\
\texttt{\textless/explanation\textgreater}
\end{minipage}%
}
\caption{Self-reflection prompt templates used by \MethodName.
The system prompt provides task-level instructions to the reflector.
In the prompt templates, \textcolor{blue}{\texttt{\{problem\}}} is replaced with the problem $x$;
\textcolor{blue}{\texttt{\{correct\_rollout\}}} is replaced with the selected correct rollout, i.e., $y^{*}$ in Equation~\ref{eq:wrong_reflection} and $y^{+}$ in Equation~\ref{eq:correct_reflection};
\textcolor{blue}{\texttt{\{wrong\_rollout\}}} is replaced with the wrong rollout $y^{-}$;
and \textcolor{blue}{\texttt{\{answer\}}} is replaced with the correct final answer.
The outputs enclosed by \texttt{\textless error\_quote\textgreater} and \texttt{\textless explanation\textgreater} correspond to the error quote $q$ and the key corrective idea $e$, respectively.}
\vspace{-10pt}
\label{fig:reflection_prompt}
\end{figure*}

\section{Token-Level KL Divergence: Valid Prefixes vs. Error Sentences}
\label{app:token_kl_analysis}

In Section~\ref{sec:introduction} and Figure~\ref{fig:overview}(b), we report that standard On-Policy Self-Distillation (SDPO) assigns unexpectedly large KL divergence supervision to valid reasoning prefixes---frequently exceeding the supervision exerted on the actual error sentence itself.
Here, we present the comprehensive empirical methodology and benchmark-wide statistics.

To investigate how self-distillation distributes supervision across sound and erroneous reasoning steps, we sample 50 questions from each of three scientific reasoning domains in SciKnowEval~\citep{feng2024sciknoweval} (\textit{Biology}, \textit{Physics}, and \textit{Materials}) and generate reasoning rollouts using the base Qwen3-4B model at training Step~0.
For each incorrect rollout paired with a verified reference solution, we employ \texttt{claude-opus-5} as an impartial adjudicator to identify the exact first erroneous sentence (\texttt{<error\_quote>}) and explain the underlying reasoning flaw (see Figure~\ref{fig:judge_error_quote_prompt} for the prompt template).
Using the extracted error span, we partition each trajectory into the valid prefix prior to the error and the subsequent error sentence, and compute the token-level KL divergence along the rollout.
As summarized in Figure~\ref{fig:overview}(b), across all three domains, standard SDPO exhibits counter-intuitive supervisory behavior: the average token KL on valid prefixes is $0.9786$ on Biology, $0.9080$ on Physics, and $0.9054$ on Materials, which consistently exceeds the average KL on the actual error sentence ($0.8881$ on Biology, $0.7605$ on Physics, and $0.8513$ on Materials).
Because tokens preceding the error are already scientifically sound, this strong pull toward the reference teacher merely penalizes harmless stylistic deviations and legitimate alternative derivations rather than rectifying flaws, directly evidencing that standard OPSD causes premature mode collapse by over-supervising valid reasoning prefixes.

\begin{figure*}[t]
\centering
\footnotesize
\fcolorbox{black}{gray!4}{
\begin{minipage}{0.96\linewidth}
\raggedright
\texttt{[System]}\\
You are an expert academic evaluator analyzing reasoning chains of language models. Your task is to pinpoint the exact location where an incorrect reasoning rollout first makes an error. You must respond strictly in the requested XML format and nothing else.

\medskip
\texttt{[User]}\\
We have a scientific reasoning problem, the ground-truth answer, a verified correct reference solution, and an incorrect student rollout.

\smallskip
Please analyze the incorrect rollout and identify the EXACT first sentence or clause where the reasoning goes wrong. \\
Respond strictly in the following format:\\
\texttt{\textless error\_quote\textgreater}\\
(Extract the EXACT substring from the [Incorrect Solution] where the reasoning first commits a factual, conceptual, or deductive error. Do not quote correct setup sentences preceding the mistake.)\\
\texttt{\textless/error\_quote\textgreater}\\
\texttt{\textless explanation\textgreater}\\
(Explain concisely what the error is and why it deviates from the correct scientific logic.)\\
\texttt{\textless/explanation\textgreater}

\medskip
\textbf{[Problem]}\\
\textcolor{blue}{\{problem\_text\}}

\smallskip
\textbf{[Ground-Truth Answer]}\\
\textcolor{blue}{\{ground\_truth\}}

\smallskip
\textbf{[Correct Reference Solution]}\\
\textcolor{blue}{\{reference\_correct\_solution\}}

\smallskip
\textbf{[Incorrect Student Solution]}\\
\textcolor{blue}{\{student\_wrong\_response\}}
\end{minipage}
}
\caption{Prompt template provided to \texttt{claude-opus-5} for extracting the exact first error quote and diagnostic explanation.}
\label{fig:judge_error_quote_prompt}
\end{figure*}

\section{Case Study: Protein Mutation Fitness Prediction}
\label{app:case_study_biology}

To supplement the qualitative case study presented in Figure~\ref{fig:overview}(c), this section provides the complete verbatim problem statement, student rollout text, and self-reflector diagnostic output for the representative diagnostic example from the Biology domain.

\paragraph{Problem Description.}
The student policy is queried to predict the functional fitness score of an experimentally mutated protein sequence (a four-site random mutant of the streptococcal protein G B1 domain):
\begin{quote}
    \small
    \textbf{Problem:} \textit{``Given the wide-type protein sequence `MQYKLILNGKTLKGETTTEAVDAATAEKVFKQYANDNGVDGEWTYDDATKTFTVTE' with an initial fitness of 1.0, we have performed a four-sites random mutation resulting in the mutant `MQYKLILNGKTLKGETTTEAVDAATAEKVFKQYANDNGMRKEWTYDDATKTFTGTE'. Here the fitness is determined by both stability (i.e. the fraction of folded proteins) and function (i.e. binding affinity to IgG-Fc receptors). Can you calculate the fitness of this mutated version?''} \\
    \textbf{Options:} \textit{A: 0.0 \quad B: 0.008 \quad C: 0.691 \quad D: 1.002} \\
    \textbf{Ground Truth:} \textbf{B} (\textit{0.008}).
\end{quote}

\paragraph{Student Reasoning Rollout.}
The sampled on-policy rollout from Qwen3-4B unfolds across three segments:
\begin{enumerate}
    \item \textit{Valid Prefix:} The model correctly recognizes the problem setup and biological background:
    \begin{quote}
        \small
        \textit{``The question provides a protein sequence and its initial fitness of 1.0. A four-sites random mutation has been performed, resulting in a mutant sequence. The fitness of the mutant is determined by both stability (fraction of folded proteins) and function (binding affinity to IgG-Fc receptors). However, the question does not provide any specific data or calculations to determine the exact fitness of the mutant. The options given (0.0, 0.008, 0.691, 1.002) suggest that the fitness has changed from the initial value of 1.0, but without additional information such as the effect of the mutation on stability or function, it is not possible to calculate the exact fitness. Given that the question asks to ``calculate the fitness'' of the mutated version, and no specific data or formula is provided,''}
    \end{quote}
    \item \textit{Error Sentence (highlighted by the red rectangle in Figure~\ref{fig:overview}(c)):} The model introduces an unfounded heuristic jump, falsely equating a fitness reduction with total loss of function:
    \begin{quote}
        \small
        \textit{``the most reasonable assumption is that the fitness has decreased due to the mutation, and the correct answer is the one with the lowest value, which is 0.0.''}
    \end{quote}
    \item \textit{Post-Error Suffix:} The model rationalizes this arbitrary assumption and concludes with the incorrect option \textbf{A}:
    \begin{quote}
        \small
        \textit{``This is a common outcome when a mutation disrupts the function or stability of a protein, leading to a significant decrease in fitness. Therefore, the correct answer is A: 0.0.''}
    \end{quote}
\end{enumerate}

\paragraph{Self-Reflector Diagnosis.}
Conditioned on the problem and the verified correct answer, our self-reflector successfully pinpoints the reasoning failure and produces the following structured reflection:
\begin{quote}
    \small
    \texttt{\textless error\_quote\textgreater}\\
    \textit{the most reasonable assumption is that the fitness has decreased due to the mutation, and the correct answer is the one with the lowest value, which is 0.0.}\\
    \texttt{\textless/error\_quote\textgreater}\\
    \texttt{\textless explanation\textgreater}\\
    \textit{The mistake is in arbitrarily assuming that the correct answer must be the lowest possible value (0.0) representing complete loss of function. While random mutations disrupt stability and function, a quadruple mutant can retain low non-zero fitness rather than complete functional collapse. Therefore, the correct fitness score corresponding to this mutant is 0.008 (Option B).}\\
    \texttt{\textless/explanation\textgreater}
\end{quote}

As discussed in Section~\ref{sec:introduction}, Figure~\ref{fig:overview}(c) visualizes the token-by-token KL divergence under both frameworks across this exact trajectory, directly demonstrating that standard SDPO penalizes sound prefixes while failing to penalize errors, whereas \MethodName shields valid prefixes and focuses corrective pressure squarely on the diagnosed logical flaw.

\section{Cross-Domain Generalization Results}
\label{app:cross_domain_results}

\begin{table*}[t]
\centering
\small
\setlength{\tabcolsep}{5pt}
\begin{tabular}{@{}llcccccc@{}}
\toprule
\rowcolor{Gainsboro} \textbf{Train on Biology} & \textbf{Method} & AIME2024 & Material & Physics & Chemistry & ToolUse & Average \\
\midrule
\multirow{3}{*}{Qwen3-4B}
& GRPO & 31.25 & \textbf{56.16} & \textbf{56.35} & \textbf{45.23} & \textbf{58.85} & \textbf{49.57} \\
& SDPO & \textbf{33.33} & 38.16 & 45.68 & 39.35 & \underline{57.05} & 42.71 \\
& \MethodName & \underline{31.94} & \underline{51.37} & \underline{49.58} & \underline{40.93} & 56.46 & \underline{46.06} \\
\cmidrule(lr){1-8}
\multirow{3}{*}{Qwen3-8B}
& GRPO & \textbf{38.89} & \textbf{59.04} & \textbf{60.89} & \textbf{41.58} & 59.16 & \textbf{51.91} \\
& SDPO & 31.25 & 38.10 & 40.36 & 34.78 & \textbf{59.44} & 40.79 \\
& \MethodName & \underline{32.64} & \underline{56.01} & \underline{56.98} & \underline{38.21} & \underline{59.38} & \underline{48.64} \\
\midrule
\rowcolor{Gainsboro} \textbf{Train on Chemistry} & \textbf{Method} & AIME2024 & Biology & Material & Physics & ToolUse & Average \\
\midrule
\multirow{3}{*}{Qwen3-4B}
& GRPO & \textbf{35.42} & 28.88 & \textbf{49.78} & \textbf{49.71} & \textbf{58.92} & \textbf{44.54} \\
& SDPO & 0.00 & \textbf{31.54} & 20.41 & 23.05 & 39.64 & 22.93 \\
& \MethodName & \underline{30.56} & \underline{30.50} & \underline{34.64} & \underline{39.95} & \underline{57.05} & \underline{38.54} \\
\cmidrule(lr){1-8}
\multirow{3}{*}{Qwen3-8B}
& GRPO & \textbf{43.06} & \underline{36.17} & \textbf{64.18} & \textbf{61.80} & \textbf{60.23} & \textbf{53.09} \\
& SDPO & 9.03 & 28.67 & 22.89 & 29.92 & 0.83 & 18.27 \\
& \MethodName & \underline{34.03} & \textbf{37.08} & \underline{50.22} & \underline{49.09} & \underline{58.27} & \underline{45.74} \\
\midrule
\rowcolor{Gainsboro} \textbf{Train on Material} & \textbf{Method} & AIME2024 & Biology & Physics & Chemistry & ToolUse & Average \\
\midrule
\multirow{3}{*}{Qwen3-4B}
& GRPO & \textbf{38.89} & \textbf{33.63} & \textbf{60.78} & \underline{42.88} & \textbf{60.85} & \textbf{47.41} \\
& SDPO & 4.86 & 28.46 & 30.29 & 40.03 & 54.87 & 31.70 \\
& \MethodName & \underline{29.86} & \underline{31.29} & \underline{45.05} & \textbf{45.48} & \underline{55.15} & \underline{41.37} \\
\cmidrule(lr){1-8}
\multirow{3}{*}{Qwen3-8B}
& GRPO & \textbf{39.58} & 27.79 & \textbf{63.39} & \textbf{46.87} & \textbf{61.31} & \textbf{47.79} \\
& SDPO & 25.00 & \textbf{35.83} & 42.76 & \underline{44.07} & 0.86 & 29.70 \\
& \MethodName & \underline{34.03} & \underline{31.75} & \underline{55.57} & 42.12 & \underline{59.38} & \underline{44.57} \\
\midrule
\rowcolor{Gainsboro} \textbf{Train on Physics} & \textbf{Method} & AIME2024 & Biology & Material & Chemistry & ToolUse & Average \\
\midrule
\multirow{3}{*}{Qwen3-4B}
& GRPO & \textbf{35.42} & 28.71 & \textbf{71.48} & \underline{43.60} & \textbf{58.98} & \textbf{47.64} \\
& SDPO & 8.33 & \underline{32.96} & 43.62 & \textbf{43.63} & 54.01 & 36.51 \\
& \MethodName & \underline{28.47} & \textbf{33.88} & \underline{62.21} & 43.43 & \underline{57.17} & \underline{45.03} \\
\cmidrule(lr){1-8}
\multirow{3}{*}{Qwen3-8B}
& GRPO & \textbf{44.44} & 25.96 & \textbf{63.76} & \textbf{46.12} & \textbf{59.93} & \textbf{48.04} \\
& SDPO & 29.17 & \textbf{34.46} & 49.38 & \underline{44.00} & 2.27 & 31.86 \\
& \MethodName & \underline{33.33} & \underline{34.17} & \underline{60.33} & 41.62 & \underline{59.19} & \underline{45.73} \\
\midrule
\rowcolor{Gainsboro} \textbf{Train on ToolUse} & \textbf{Method} & AIME2024 & Biology & Material & Physics & Chemistry & Average \\
\midrule
\multirow{3}{*}{Qwen3-4B}
& GRPO & \textbf{33.33} & \textbf{30.54} & \underline{52.70} & \textbf{57.29} & \underline{43.97} & \textbf{43.57} \\
& SDPO & 11.11 & 0.88 & 1.09 & 0.49 & 0.83 & 2.88 \\
& \MethodName & \underline{31.94} & \underline{29.63} & \textbf{53.30} & \underline{46.25} & \textbf{45.43} & \underline{41.31} \\
\cmidrule(lr){1-8}
\multirow{3}{*}{Qwen3-8B}
& GRPO & \textbf{36.11} & \underline{36.13} & \textbf{58.82} & \textbf{57.76} & 42.47 & \textbf{46.26} \\
& SDPO & 31.94 & 35.21 & 44.22 & 40.34 & \textbf{45.18} & 39.38 \\
& \MethodName & \underline{33.33} & \textbf{37.04} & \underline{55.87} & \underline{54.30} & \underline{45.03} & \underline{45.11} \\
\bottomrule
\end{tabular}
\caption{Cross-domain generalization results.
All methods are trained to convergence on the training dataset and scores are reported as mean@16 (\%) averaged over the last three evaluation steps.
\textbf{Boldface} indicates the best result,
and \underline{underline} indicates the second-best result within each model scale.}
\label{tab:ood_results}
\end{table*}

To evaluate whether the models acquire robust, generalizable reasoning capabilities beyond their specific training domains,
we conduct cross-domain evaluation where models trained on each source domain are evaluated across all other held-out domains as well as the out-of-domain mathematical reasoning benchmark AIME2024~\citep{feng2024sciknoweval}.
Table~\ref{tab:ood_results} reports the cross-domain results across all five training datasets on Qwen3-4B and Qwen3-8B.

\paragraph{GRPO demonstrates strong cross-domain transfer.}
Across training datasets and model scales, the reinforcement learning baseline (GRPO) shows strong cross-domain transfer.
This is because GRPO relies solely on outcome-level reward verification without imposing strict token-level distribution constraints, thereby encouraging policy exploration and mitigating domain-specific overfitting.
In contrast, on-policy self-distillation introduces dense token-level KL divergence constraints against teacher trajectories, which can make the model more prone to overfitting to domain-specific phrasing.

\paragraph{\MethodName substantially alleviates distillation collapse and outperforms SDPO.}
Despite the exploration benefits of outcome-level RL, \MethodName consistently outperforms standard SDPO on cross-domain average accuracy across all training datasets and both model scales.
Standard SDPO suffers catastrophic collapse when transferring across substantially different domains: for instance, when trained on ToolUse and evaluated on scientific question-answering and mathematical reasoning, SDPO collapses to an average score of only 2.88\% on Qwen3-4B (compared to 43.57\% for GRPO).
In sharp contrast, \MethodName preserves strong reasoning transfer, maintaining 41.31\% under the same setting.
This demonstrates that by diagnosing error spans and restricting self-distillation exclusively to erroneous suffixes while safeguarding valid reasoning prefixes, \MethodName prevents the memorization of brittle domain-specific syntactic templates, retaining strong cross-domain generalization while benefiting from dense on-policy supervision.

\section{Training Efficiency}
\label{app:training_efficiency}

\begin{figure}[t]
\centering
\begin{minipage}[t]{0.35\textwidth}
    \centering
    \includegraphics[width=\linewidth]{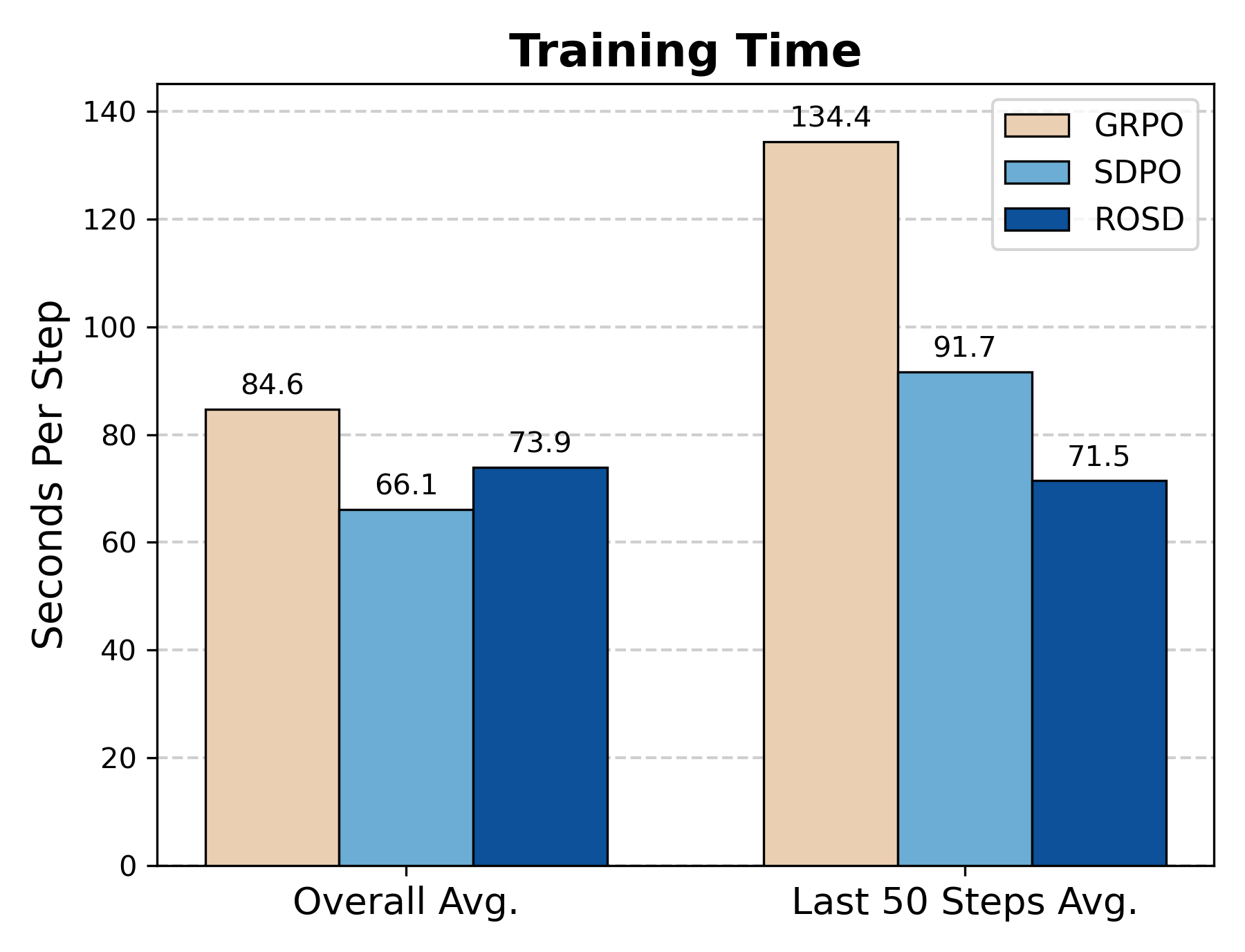}
    \caption{Training efficiency dynamics on Chemistry with Qwen3-8B across training steps (time in seconds per step for total steps and the last 50 steps).}
    \label{fig:training_time}
\end{minipage}
\hspace{0.02\textwidth}
\begin{minipage}[t]{0.35\textwidth}
    \centering
    \includegraphics[width=\linewidth]{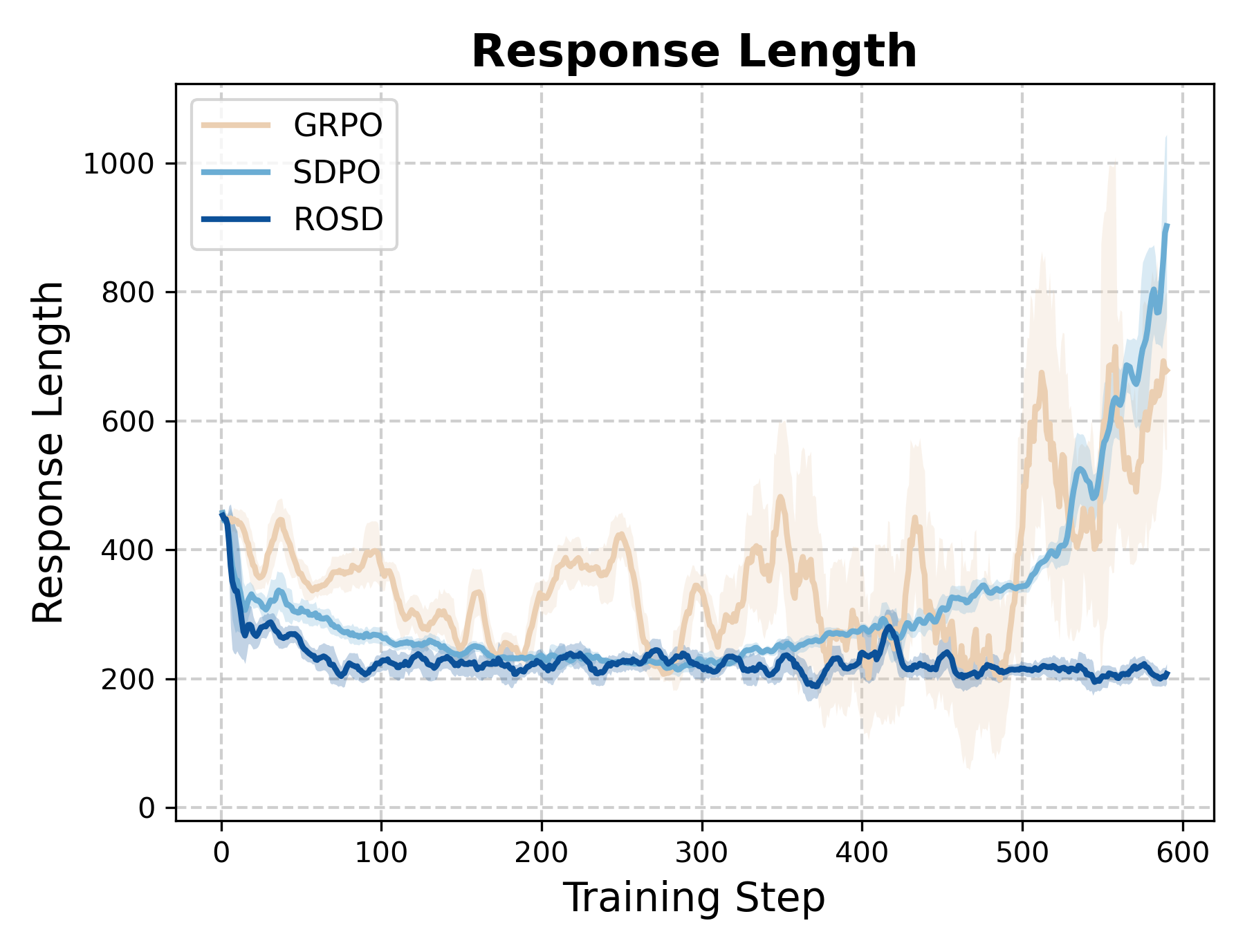}
    \caption{Response length dynamics on Chemistry with Qwen3-8B (rolling average of response length over 10 steps).}
    \label{fig:response_length}
\end{minipage}
\end{figure}

We evaluate the training efficiency dynamics on Chemistry with the representative Qwen3-8B backbone.
As shown in Figure~\ref{fig:training_time}, although \MethodName introduces an additional self-reflection step, its overall per-step wall-clock training time is comparable to standard SDPO and consistently lower than GRPO.
In later training stages, \MethodName becomes even more computationally efficient: because the response lengths of the baseline methods (GRPO and SDPO) expand significantly, they incur progressively higher attention and decoding overhead.
In contrast, because \MethodName maintains concise and stable response lengths, its training time per step remains steady throughout optimization.

\section{Response Length}
\label{app:response_length}

We evaluate the response length dynamics on Chemistry with the representative Qwen3-8B backbone.
As illustrated in Figure~\ref{fig:response_length}, \MethodName produces relatively shorter responses and maintains a stable response-length trajectory throughout training.
In contrast, the response lengths of both GRPO and SDPO increase substantially as training proceeds.
By focusing distillation strictly on the diagnosed error span without forcing full-trajectory imitation, \MethodName avoids superfluous verbose phrasing and prevents unnecessary reasoning dilation, demonstrating superior optimization stability.

\section{Error Localization Dynamics}
\label{app:error_localization_dynamics}

\begin{figure}[t]
\centering
\includegraphics[width=0.7\linewidth]{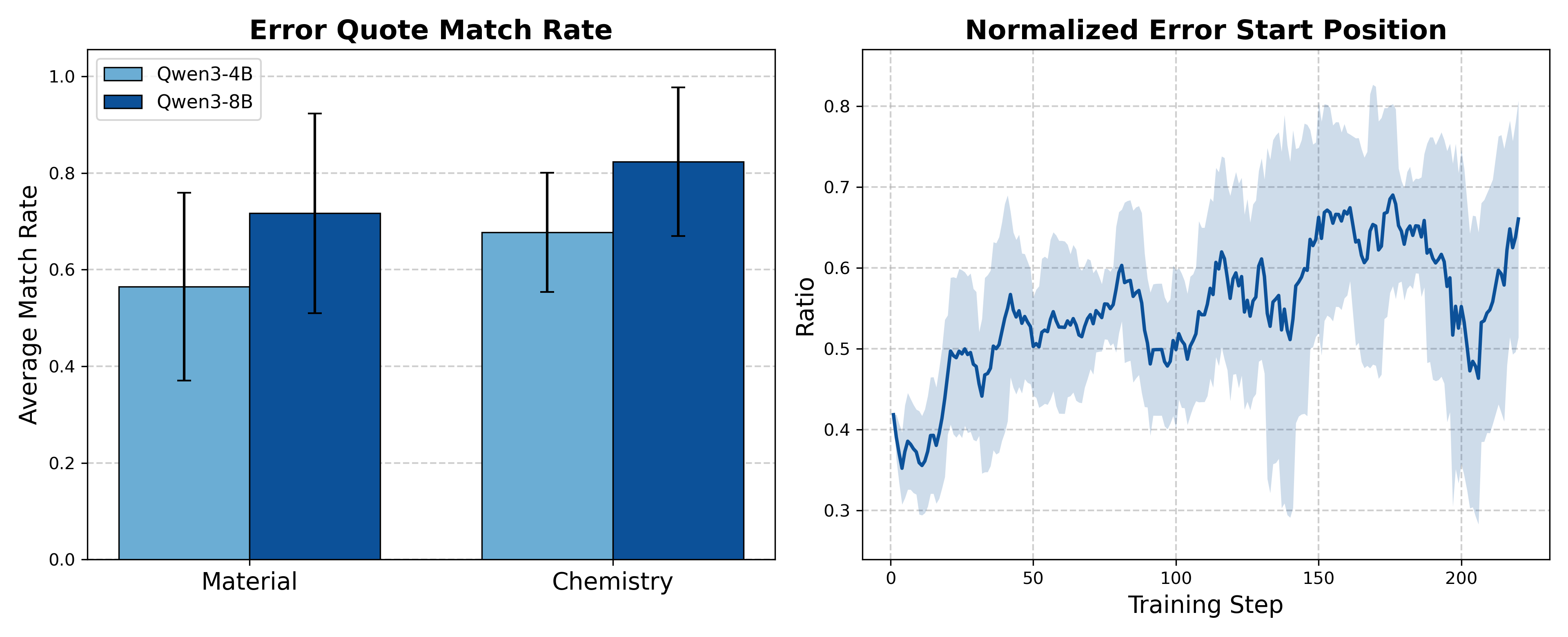}
\caption{Error localization dynamics during training.
The left figure shows the average match rate of error quotes produced by \MethodName on Material and Chemistry with Qwen3-4B and Qwen3-8B.
The right figure shows the average normalized error position over 10 steps on Physics with Qwen3-8B.}
\label{fig:discussion_dynamics}
\end{figure}

Figure~\ref{fig:discussion_dynamics} provides additional evidence for the localization behavior of \MethodName.
The left figure shows that the error quote match rate maintains relatively high values on both Material and Chemistry,
indicating that a substantial fraction of reflector outputs can be aligned back to the original rollout.
The Qwen3-8B model achieves a higher match rate than Qwen3-4B on both datasets,
which is consistent with its stronger instruction-following ability.
The right figure shows that the normalized error position gradually moves toward the later part of the response,
suggesting that errors become less frequent and occur later as training progresses.
This trend also explains why full-response distillation can become harmful in the late stage:
when most early reasoning steps are already valid,
continuing to distill the whole response may unnecessarily perturb correct prefixes.
These observations further support the necessity of localizing self-distillation to the diagnosed error span.

\section{Prompt Templates}
\label{app:prompt_templates}

Following SDPO~\citep{hubotter2026reinforcement},
we use the same dataset prompts for student rollout and evaluation.
For Qwen models,
we disable the thinking-mode chat-template option during prompt construction.
The concrete ScienceQA, ToolUse, and mathematics prompt templates are shown in
Figures~\ref{fig:scienceqa_system_prompt}--\ref{fig:math_prompt}.
The prompt used as input to the self-teacher is shown in Figure~\ref{fig:teacher_prompt_template}.

\begin{figure}[!htbp]
\centering
\footnotesize
\fcolorbox{black}{gray!4}{
\begin{minipage}{0.96\linewidth}
\raggedright
Given a question and four options, please select the right answer.
Respond in the following format:\\
\texttt{\textless reasoning\textgreater}\\
\texttt{...}\\
\texttt{\textless/reasoning\textgreater}\\
\texttt{\textless answer\textgreater}\\
\texttt{...}\\
\texttt{\textless/answer\textgreater}

\smallskip
For the answer, only output the letter corresponding to the correct option
(A, B, C, or D), and nothing else.
Do not restate the answer text.
For example, if the answer is ``A'', just output:\\
\texttt{\textless answer\textgreater}\\
\texttt{A}\\
\texttt{\textless/answer\textgreater}
\end{minipage}
}
\caption{System prompt for ScienceQA.}
\label{fig:scienceqa_system_prompt}
\end{figure}

\begin{figure}[!htbp]
\centering
\footnotesize
\fcolorbox{black}{gray!4}{
\begin{minipage}{0.96\linewidth}
\raggedright
\texttt{[User]} \textcolor{blue}{\{question\}}\\
\mbox{}\\
Please reason step by step.
\end{minipage}
}
\caption{User prompt for ScienceQA.}
\label{fig:scienceqa_user_prompt}
\end{figure}

\begin{figure}[!htbp]
\centering
\footnotesize
\fcolorbox{black}{gray!4}{
\begin{minipage}{0.96\linewidth}
\raggedright
\texttt{[User]} Your task is to answer the user's question using available tools.\\
You have access to the following tools:\\
\texttt{Name:} \textcolor{blue}{\{tool\_name\}}\\
\texttt{Description:} \textcolor{blue}{\{tool\_description\}}\\
\texttt{Documentation:}\\
\textcolor{blue}{\{function\_name\}}: \textcolor{blue}{\{function\_description\}}\\
\texttt{Parameters:} \textcolor{blue}{\{parameter\_schema\}}\\
\texttt{Output:} \textcolor{blue}{\{output\_description\}}\\
\texttt{- Format:} \textcolor{blue}{\{output\_format\}}\\
\texttt{- Structure:} \textcolor{blue}{\{output\_structure\}}\\
\mbox{}\\
Use the following format:\\
\texttt{Thought:} think about what to do\\
\texttt{Action:} the action to take; use one tool name.\\
\texttt{Action Input:} the input in JSON format.\\
\texttt{Begin!}\\
\texttt{Question:} \textcolor{blue}{\{question\}}
\end{minipage}
}
\caption{User prompt for ToolUse.}
\label{fig:tooluse_prompt}
\end{figure}

\begin{figure}[!htbp]
\centering
\footnotesize
\fcolorbox{black}{gray!4}{
\begin{minipage}{0.96\linewidth}
\raggedright
\texttt{[User]} \textcolor{blue}{\{question\}}\\
\mbox{}\\
Please reason step by step, and put your final answer within
\texttt{\textbackslash boxed\{\}}.
\end{minipage}
}
\caption{User prompt for AIME2024.}
\label{fig:math_prompt}
\end{figure}

\begin{figure}[!htbp]
\centering
\footnotesize
\fcolorbox{black}{gray!4}{
\begin{minipage}{0.96\linewidth}
\raggedright
\texttt{[System]} \textcolor{blue}{\{original system prompt\}}\\
\texttt{[User]} \textcolor{blue}{\{original user prompt\}}\\
The following is the key idea to solve the question:\\
\textcolor{blue}{\{key\_idea\}}\\
Correctly solve the original question. \\
\texttt{[Assistant]} \textcolor{blue}{\{original rollout\}}\\
\end{minipage}
}
\caption[Teacher input template with reflection.]{Teacher input template with reflection.
The self-teacher preserves the original system and user prompts used for rollout generation.
The placeholder \texttt{\{key\_idea\}} is replaced with the key idea generated by the self-reflector.
}
\label{fig:teacher_prompt_template}
\end{figure}

\end{document}